\documentclass[10pt,letterpaper]{article}
\usepackage{amsmath,amssymb,graphicx,hyperref,amsfonts,bm}
\usepackage{caption,subcaption}
\usepackage{booktabs}
\usepackage[super,square,comma,sort&compress]{natbib}
\usepackage{authblk}
\usepackage{tikz}
\usetikzlibrary{arrows.meta}
\usetikzlibrary{positioning}
\usetikzlibrary{shapes.geometric}
\usetikzlibrary{calc}
\usepackage{float}
\usepackage[dvipsnames]{xcolor}

\definecolor{classical}{RGB}{220,80,70}
\definecolor{quantum}{RGB}{240,180,60}
\definecolor{vqc}{RGB}{130,190,110}
\definecolor{measure}{RGB}{130,160,210}
\definecolor{fusion}{RGB}{160,120,200}
\definecolor{classifier}{RGB}{70,70,70}
\definecolor{myTeal}{RGB}{0,128,128}

\usepackage[T1]{fontenc}
\usepackage[utf8]{inputenc}
\usepackage{lmodern}
\usepackage{microtype}

\hypersetup{
  colorlinks=true,
  linkcolor=blue,
  citecolor=blue,
  urlcolor=blue
}

\usepackage[margin=1in]{geometry}
\usepackage{siunitx}
\title{Practical Quantum-Classical Feature Fusion for complex data Classification}

\author[1]{Azadeh Alavi*}
\author[2]{Fatemeh Kouchmeshki}
\author[4]{Abdolrahman Alavi}
\affil[1]{RMIT University, Melbourne, Australia}
\affil[1-4]{Pattern Recognition Pty. Ltd. Melbourne, Asutralia \\ Email azadeh@pr2aid.com*}

\date{}

\begin{document}

\maketitle
\begingroup
\renewcommand\thefootnote{}\footnotetext{First and second authors contributed equally to this work.}
\addtocounter{footnote}{0}
\endgroup

\begin{abstract}
\begin{abstract}
Hybrid quantum and classical learning aims to couple quantum feature maps with the robustness of classical neural networks, yet most architectures treat the quantum circuit as an isolated feature extractor and merge its measurements with classical representations by direct concatenation. This neglects that the quantum and classical branches constitute distinct computational modalities and limits reliable performance on complex, high dimensional tabular and semi structured data, including remote sensing, environmental monitoring, and medical diagnostics. We present a multimodal formulation of hybrid learning and propose a cross attention mid fusion architecture in which a classical representation queries quantum derived feature tokens through an attention block with residual connectivity. The quantum branch is kept within practical NISQ budgets and uses up to nine qubits. We evaluate on Wine, Breast Cancer, Forest CoverType, FashionMNIST, and SteelPlatesFaults, comparing a quantum only model, a classical baseline, residual hybrid models, and the proposed mid fusion model under a consistent protocol. Pure quantum and standard hybrid designs underperform due to measurement induced information loss, while cross attention mid fusion is consistently competitive and improves performance on the more complex datasets in most cases. These findings suggest that quantum derived information becomes most valuable when integrated through principled multimodal fusion rather than used in isolation or loosely appended to classical features.
\end{abstract}

\end{abstract}


\section{Introduction}

Hybrid quantum and classical learning has attracted increasing interest as a strategy for enriching machine learning pipelines with quantum feature maps while retaining the robustness and scalability of classical models. Machine learning itself has enabled automation and scientific progress across many disciplines, ranging from industrial process monitoring and medical image screening to the analysis of large scale sensor streams. Classical approaches, including linear estimators, kernel methods, and modern deep neural networks, often deliver strong performance. However, as dimensionality increases and statistical dependencies become more tightly coupled, computationally tractable feature transformations frequently compress, smooth, and approximate in ways that suppress the structure most relevant to downstream decisions.

Quantum information processing offers a distinct set of computational primitives. By exploiting superposition and entanglement, quantum circuits operate on Hilbert spaces whose dimension grows exponentially with the number of qubits \cite{nielsen_chuang_book}. This motivates the use of quantum circuits as expressive representations that are challenging to emulate efficiently with classical resources. At the same time, current devices remain in the noisy intermediate scale quantum regime, where qubit counts are limited, gates are imperfect, and decoherence imposes persistent constraints \cite{DeLuca2022Survey}. These realities have focused attention on hybrid quantum and classical architectures, in which a variational quantum circuit is trained jointly with a classical model.

Most of the hybrid quantum and classical literature follows a modular pipeline: classical data are encoded into a parameterized circuit, expectation values are measured, and the resulting measurement features are supplied to a classical predictor \cite{Tomar2025_QMLSurvey,PQCs2019_Benedetti}. This design treats the quantum circuit primarily as a nonlinear feature map, aligning with the broader variational paradigm that also underlies algorithms such as the variational quantum eigensolver \cite{VQE_Algorithm}. Once both quantum and classical representations are available, the key architectural question concerns the mechanism used to integrate them into a coherent predictor.

A common integration choice is direct concatenation of classical features and quantum measurement features. While effective as a baseline, concatenation often defers cross stream interaction to a shallow classical head, limiting the extent to which complementary structure is exploited. Recent hybrid designs indicate that more explicit pathways for quantum information, including residual and bypass style connections, can improve stability and make the quantum signal easier to use \cite{Zhang2025ReadoutBypass_a}. In parallel, multimodal learning has established that effective performance depends on deliberate fusion choices, including the stage at which fusion occurs and the mechanism that regulates how strongly each modality influences the other \cite{Baltruaitis2018MultimodalReview_a}.

This work formulates hybrid quantum and classical classification as a multimodal learning problem, in which a classical network and a variational quantum circuit are treated as two computational modalities that process the same input under distinct inductive biases. We preserve modality specific representations and study fusion as an explicit design axis. Under a shared preprocessing and evaluation protocol, we compare families of early, mid, deep, and late integration architectures. Mid fusion is implemented in two ways: a gated latent mixing module and an attention based block that treats quantum readouts as a short sequence of tokens and refines a classical CLS representation through Transformer style self attention \cite{vaswani2017attention}. To reflect realistic resource budgets and to keep repeated simulation tractable, we cap quantum width at $Q\in\{7,8,9\}$ qubits and sweep circuit depth over $L\in\{1,2,3\}$ layers of a hardware efficient ansatz, with optional data reuploading \cite{perezsalinas2020reupload}. Empirical evaluation is conducted on five fixed length vector classification tasks spanning low dimensional and high dimensional regimes: Wine, Breast Cancer Wisconsin (Diagnostic), a three class subset of Covertype, a three class subsample of Fashion MNIST flattened to $\mathbb{R}^{784}$, and Steel Plates Faults.

\noindent\textbf{Contributions.} The main contributions of this work are:
\begin{itemize}
    \item \textbf{Multimodal formulation of hybrid learning:} We cast hybrid quantum and classical classification as a multimodal learning problem, treating classical latent features and quantum measurement features as distinct modalities that capture different relationships and dependencies in the data.

    \item \textbf{Cross attention mid fusion for modality interaction:} We introduce a mid fusion architecture in which a classical representation queries quantum derived feature tokens via attention with residual connections. This preserves modality specific structure while enabling sample adaptive integration of quantum information.

    \item \textbf{Controlled evaluation across datasets:} We benchmark on Wine, Wisconsin Breast Cancer, Fashion MNIST, and Forest CoverType using five fold cross validation, reporting accuracy, precision, recall, and F1 score under a consistent training protocol.

    \item \textbf{Measured gains over Residual 6q Deep:} Compared with the Residual 6q Deep baseline, cross attention mid fusion improves mean accuracy by 3.4 percentage points on Wine, 1.5 on Wisconsin Breast Cancer, 9.2 on Fashion MNIST, and 4.4 on Forest CoverType. In the same setting, classical only and quantum only baselines reach 72.9\% to 96.1\% and 36.5\% to 62.7\% accuracy, respectively, underscoring that performance improvements follow from structured multimodal fusion.
\end{itemize}

The rest of the paper proceeds as follows. Section~\ref{sec:background} reviews the relevant background on QML, HQC
architectures, and fusion mechanisms. Section~\ref{sec:methodology} details our two-stream construction, preprocessing,
quantum layer design, and the fusion families considered. Section~\ref{sec:results} describes the experimental
setup and reports results, and discusses implications and limitations. We conclude in
Section~\ref{sec:conclusion}.

\section{Background and Related Work}
\label{sec:background}

Quantum machine learning (QML) has developed into a fast-moving research area at the interface of quantum information theory and classical statistical learning.  The familiar motivation is that quantum-mechanical effects, superposition, entanglement, and interference can support transformations that are difficult, or in some regimes implausible, for classical systems to reproduce efficiently \cite{nielsen_chuang_book,Biamonte2017QML}.  Yet large-scale fault-tolerant quantum computers remain out of reach.  What we can program today largely sits in the noisy intermediate-scale quantum (NISQ) regime, where limited qubit counts and imperfect gates constrain circuit depth and reliability \cite{Preskill2018quantumcomputingin}.  This has steered the community toward hybrid quantum--classical (HQC) methods, which make selective use of quantum circuits while retaining classical control and optimisation \cite{DeLuca2022Survey}.  In that sense, HQC is less a compromise than a practical laboratory: it lets us study quantum-enabled learning systems under realistic engineering constraints.

Early work in QML explored ``fully quantum'' learning models, such as: quantum perceptrons, quantum Boltzmann machines, quantum support vector machines, and related constructions, mostly as conceptual demonstrations and as probes of what a quantum learner might represent \cite{Biamonte2017QML}.  As hardware limitations became unavoidable, research momentum shifted toward parameterized quantum circuits (PQCs) and variational quantum algorithms (VQAs), which balance expressivity against implementability on near-term devices \cite{PQCs2019_Benedetti,Tomar2025_QMLSurvey}.  A common recipe is now well established: classical inputs are embedded into quantum states (for example via angle, amplitude, or basis encodings), the state is transformed by layers of trainable unitaries built from a native gate set, and measurement yields a small set of expectation values that can be treated as nonlinear features \cite{PQCs2019_Benedetti}.  This viewpoint connects naturally to kernel methods and feature maps, where the quantum device induces an implicit high-dimensional representation that a classical learner can exploit \cite{Schuld2019featureHilbert}.

VQAs provide the optimisation template behind many PQC-based models.  The variational quantum eigensolver (VQE) is the canonical instance from quantum simulation, demonstrating a closed-loop workflow in which a classical optimizer updates circuit parameters to minimise a cost evaluated on a quantum processor \cite{VQE_Algorithm}.  While VQE was not designed for supervised learning, it crystallised a pattern that QML has repeatedly reused: train a parameterised circuit by differentiating through expectation values and iterating with classical optimisation \cite{cerezo2021vqa}.  The pattern is powerful, but it is also temperamental.  One recurrent difficulty is trainability: as qubit count and circuit depth grow, optimisation landscapes can exhibit barren plateaus regions where gradients concentrate near zero and training becomes effectively stalled \cite{McClean2018Barren}.  This has motivated substantial work on the expressivity and geometry of PQCs, the role of entanglement, and the interaction between data structure and the chosen embedding.  In parallel, tools from information geometry have been adapted to the quantum setting, including quantum natural-gradient methods that exploit the quantum geometric tensor to precondition optimisation updates \cite{Stokes2020quantumnatural}.  These geometric views do not eliminate NISQ constraints, but they have helped clarify which circuit families are plausibly trainable and how classical preprocessing and ansatz selection affect stability \cite{cerezo2021vqa}.

More recent efforts have focused on where, exactly, a quantum circuit should sit inside a modern learning pipeline.  Rather than replacing an entire model with a PQC, many hybrids use quantum components as targeted modules: a feature extractor, a regularising transformation, or a small ``head'' operating on compact latents.  Quantum transfer learning is a representative example, augmenting a pre-trained classical backbone with a trainable variational quantum layer and demonstrating the practicality of this pattern on image-recognition benchmarks \cite{Mari2020transferlearningin}.  Other proposals pursue the complementary direction, inserting quantum transformations earlier in the pipeline as local feature maps (for example, through quanvolutional layers) before passing features to a classical network \cite{Henderson2019quanvolution}.  Collectively, these systems suggest that HQC models behave most reliably when the quantum component is used where its inductive bias is plausible and its resource demands remain bounded, while classical modules handle the heavy lifting of representation learning at scale.

A particularly relevant recent contribution is the work of Zhang et al.\ \cite{Zhang2025ReadoutBypass_a}, who examined hybrid models combining classical embeddings with quantum-generated features.  Their approach treats the quantum circuit primarily as a feature-mapping mechanism: the resulting quantum outputs are concatenated with classical features and then passed to a classifier.  This ``append-and-classify'' strategy is simple and often robust, but it also largely treats the two streams as independent, leaving little room for structured interaction.  In most existing HQC literature, this is the default: fusion is implemented as direct concatenation, occasionally with a residual bypass, but rarely with an explicit model of how quantum and classical transformations might complement one another.

Our work departs from that default by explicitly treating classical and quantum representations as separate but related modalities.  This distinction is more than terminological.  It allows us to draw from the multimodal learning literature, where the question of \emph{how} to fuse information streams is recognised as central, and where fusion mechanisms are designed to support alignment, selective emphasis, and context-dependent interaction \cite{Baltruaitis2018MultimodalReview_a}.  In particular, we introduce cross-attention based mid-fusion alongside early- and late-fusion baselines, so that the classical and quantum streams can condition on one another during representation learning \cite{vaswani2017attention}.  This provides a more principled framework for understanding when quantum features add value and how they should be integrated with classical processing in practice.

\begin{figure}[H]
\centering
\begin{tikzpicture}[
    node distance=2.0cm,
    every node/.style={font=\footnotesize},
    enc/.style={
        rectangle, rounded corners,
        draw,
        minimum width=3.0cm,
        minimum height=1.0cm,
        align=center
    },
    arrow/.style={->, thick}
]

\definecolor{classical}{RGB}{220,80,70}
\definecolor{quantum}{RGB}{240,180,60}
\definecolor{vqc}{RGB}{130,190,110}
\definecolor{measure}{RGB}{130,160,210}
\definecolor{fusion}{RGB}{160,120,200}
\definecolor{classifier}{RGB}{70,70,70}


\node[enc, draw=classical] (input) {Input $\mathbf{x}$};


\node[enc, draw=classical, below=1.4cm of input] (clenc)
{Classical Encoder\\ $\displaystyle \mathbf{h}_{\mathrm{cl}} = f_{\theta}(\mathbf{x})$};

\draw[arrow] (input.south) -- (clenc.north);


\node[enc, draw=fusion, below=1.6cm of clenc] (fusionblock)
{Fusion Block\\ (Early / Late / Cross-Attention)};

\draw[arrow] (clenc.south) -- (fusionblock.north);


\node[enc, draw=quantum, right=5.2cm of input] (embed)
{Quantum Embedding\\ $\displaystyle |\psi_{0}(\mathbf{x})\rangle$};

\node[enc, draw=vqc, below=2.0cm of embed] (vqc1)
{Variational Layer 1\\ $\displaystyle U_{\phi}^{(1)}$};

\node[enc, draw=vqc, below=1.6cm of vqc1] (vqc2)
{Variational Layer 2\\ $\displaystyle U_{\phi}^{(2)}$};

\node[enc, draw=measure, below=1.6cm of vqc2] (meas)
{Measurement\\ $\displaystyle \mathbf{h}_{\mathrm{qm}}$};

\draw[arrow] (input.east) -- (embed.west);
\draw[arrow] (embed) -- (vqc1);
\draw[arrow] (vqc1) -- (vqc2);
\draw[arrow] (vqc2) -- (meas);


\draw[arrow] 
(meas.west) -- ++(-2.3cm,0) |- (fusionblock.east);


\node[enc, draw=classifier, below=2.0cm of fusionblock] (clf)
{Classifier \\ $\displaystyle g_{\psi}$};

\node[enc, draw=classifier, below=1.5cm of clf] (out)
{Prediction $y$};

\draw[arrow] (fusionblock.south) -- (clf.north);
\draw[arrow] (clf.south) -- (out.north);

\end{tikzpicture}
\caption{
Hybrid quantum--classical architecture studied in this work.  Classical and quantum encoders produce complementary latent representations, which are merged via fusion strategies (early, late, or cross-attention) before classification.}
\label{fig:full-architecture}
\end{figure}
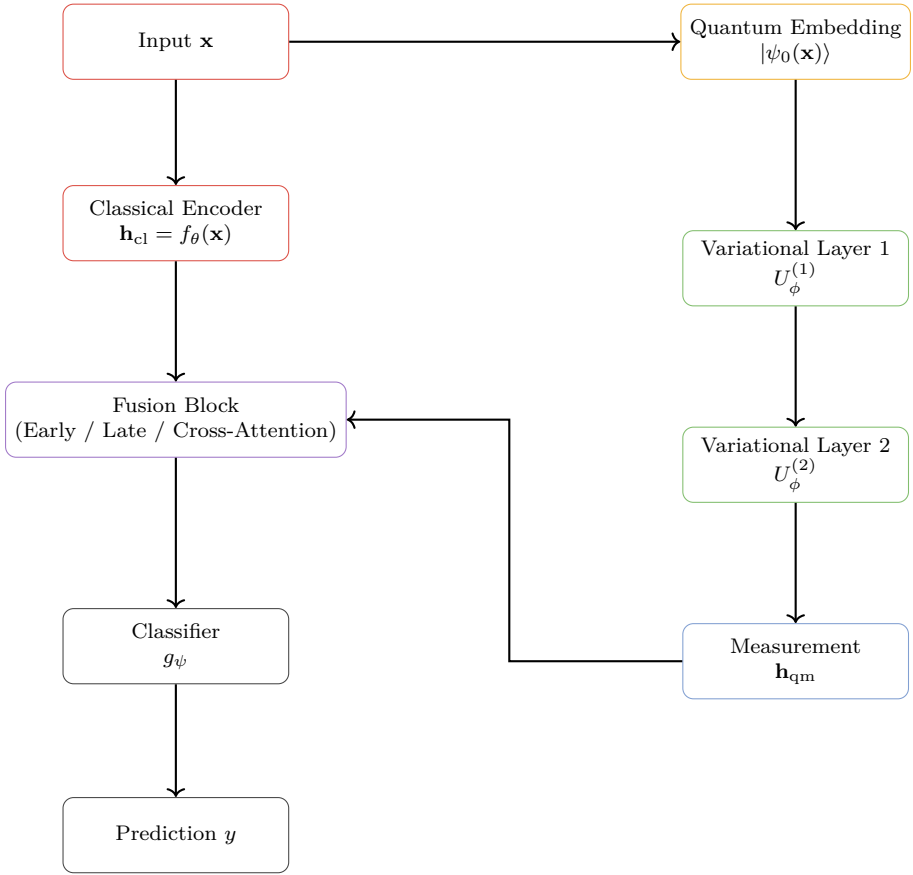

In parallel, software ecosystems such as PennyLane have made HQC experimentation markedly more accessible.  PennyLane provides differentiable quantum simulators, automatic differentiation through quantum nodes, and consistent interfaces to hardware backends, which together support rapid prototyping of circuits and hybrid training loops \cite{bergholm2018pennylane,PennyLane}.  This has broadened participation in QML and made systematic comparisons across architectures easier to run though it has also raised the bar for careful benchmarking, since small changes in preprocessing, optimisation, or hyperparameters can dominate the behaviour of NISQ-scale models.

Multimodal learning, although developed primarily within classical deep learning, offers particularly useful guidance here.  In multimodal architectures, modalities such as text, images, speech, or sensor streams contribute distinct but complementary views of the underlying phenomenon, and fusion strategies (early, late, or intermediate) determine how these views interact \cite{Baltruaitis2018MultimodalReview_a}.  Early fusion operates at the feature level, whereas late fusion aggregates predictions.  Intermediate fusion, especially through attention, has emerged as one of the most expressive approaches because it allows one stream to refine its representation by selectively attending to another \cite{vaswani2017attention}.

Although the quantum and classical features in our setting originate from the same dataset, their transformation processes differ in a way that is meaningfully analogous to distinct modalities.  The quantum encoder applies a unitary transformation in a high-dimensional Hilbert space and can mix correlations through entangling operations before measurement compresses the result into a small number of statistics.  By contrast, classical encoders, multilayer perceptrons, convolutional networks, and their variants, build representations through cascades of affine maps and pointwise nonlinearities, shaped by architectural priors such as locality and hierarchical composition.  Treating these pathways as modalities gives a natural justification for importing multimodal fusion: simple concatenation assumes the streams are independent feature vectors, while structured fusion can learn when quantum features contribute genuinely complementary information and when they should be down-weighted or ignored.

A handful of HQC studies have begun to explore such ideas implicitly through residual hybrids, weighted feature mixers, and gated integration mechanisms \cite{Zhang2025ReadoutBypass_a}.  Explicit cross-attention and principled multimodal fusion, however, remain comparatively underexplored in hybrid quantum modelling.  By making the fusion mechanism a first-class design choice, and by benchmarking multiple fusion strategies under a consistent training protocol, the present work aims to clarify not only empirical performance but also the conceptual foundations of hybrid architectures: when quantum and classical representations are complementary, how interaction can be learned, and where quantum components plausibly belong in end-to-end learning systems.

\section{Methodology}
\label{sec:methodology}

Our viewpoint differs from the existing ``hybrid'' QML literature:
we treat the classical network and the variational quantum circuit (VQC) as two \emph{computational
modalities}. Even though both are exposed to the same input instance, classical and quantum processing
are governed by fundamentally different inductive biases, affine maps composed with pointwise
nonlinearities on the one hand, and unitary evolution plus measurement on the other. Rather than
forcing one representation to dominate, we preserve both and study how they should be fused to form a
more informative representation for downstream classification.

To that end, we cast fusion as a design axis and compare a family of early-, mid-, deep-, and
late-integration architectures under a shared preprocessing, optimization, and evaluation protocol.
The quantum width is capped at $Q\in\{7,8,9\}$ qubits and the circuit depth is swept over
$L\in\{1,2,3\}$ layers of a hardware-efficient variational ansatz, reflecting resource budgets that are
realistic for contemporary NISQ-era devices and for repeated simulation during benchmarking.

\subsection{Datasets and learning tasks}
\label{sec:method_datasets}

We study supervised classification from fixed-length real-valued inputs. For a given dataset we write
$\mathcal{D}=\{(\mathbf{x}_{i},y_{i})\}_{i=1}^{N}$, where $\mathbf{x}_{i}\in\mathbb{R}^{d}$ is the feature vector,
$y_{i}\in\{0,\ldots,C-1\}$ is the class label, $N$ is the number of instances, $d$ is the raw feature dimension,
and $C$ is the number of classes. All features are cast to \texttt{float32}, and labels are remapped to contiguous
integers $\{0,\ldots,C-1\}$ for convenience. Table~\ref{tab:datasets} summarizes the benchmark suite used in our
experiments.
\footnote{Wine and Breast Cancer Wisconsin (Diagnostic) are distributed through \textsc{scikit-learn}'s dataset module \cite{pedregosa2011scikit}.
The remaining datasets are publicly available through OpenML \cite{vanschoren2013openml} and the UCI Machine Learning Repository \cite{dua2019uci};
our released code records the corresponding dataset identifiers and sampling seeds to ensure exact reproducibility.}

\begin{table}[t]
\centering
\caption{Summary of benchmark datasets. For the largest datasets we restrict to a three-class subset and draw
a stratified subsample to keep repeated quantum simulations tractable.}
\label{tab:datasets}
\small
\begin{tabular}{lccc}
\hline
Dataset & $N$ & $d$ & $C$ \\
\hline
Wine \cite{uci_wine,aeberhard1994wine} & 178 & 13 & 3 \\
Breast Cancer Wisconsin (Diagnostic) \cite{uci_wdbc,street1993nuclear} & 569 & 30 & 2 \\
Covertype (classes 1--3; subsampled) \cite{uci_covertype,blackard1999covertype} & $\le 5000$ & 54 & 3 \\
Fashion-MNIST (classes 0--2; subsampled) \cite{xiao2017fashion} & $\le 3000$ & 784 & 3 \\
Steel Plates Faults \cite{uci_steel} & 1941 & 27 & 7 \\
\hline
\end{tabular}
\end{table}

\paragraph{Wine.}
The Wine benchmark contains $N=178$ chemical analyses of wines grown in the same Italian region but
derived from three different cultivars ($C=3$), with $d=13$ measured constituents per sample
\cite{uci_wine,aeberhard1994wine}. We retain the standard three-class labeling.

\paragraph{Breast Cancer Wisconsin (Diagnostic).}
The WDBC dataset contains $N=569$ samples with $d=30$ real-valued features computed from digitized
fine-needle aspirate images of breast masses and a binary diagnosis label ($C=2$)
\cite{uci_wdbc,street1993nuclear}. We do not perform task-specific feature selection.

\paragraph{Covertype.}
Covertype is a large-scale cartographic dataset for forest cover type prediction, originally comprising
$581{,}012$ instances with $d=54$ features and $C=7$ cover types \cite{uci_covertype,blackard1999covertype}.
To keep repeated quantum simulations tractable and to focus the comparison on fusion mechanisms, we
restrict to classes $\{1,2,3\}$ (as labeled in the repository), draw a seeded stratified subsample of at most
$N=5000$ instances, and remap labels to $\{0,1,2\}$.

\paragraph{Fashion-MNIST.}
Fashion-MNIST is a $28\times 28$ grayscale image dataset designed as a drop-in replacement for MNIST
\cite{xiao2017fashion}. We treat each image as a flattened vector in $\mathbb{R}^{784}$, restrict to
classes $\{0,1,2\}$, and draw a seeded stratified subsample of at most $N=3000$ instances. The resulting task is
deliberately high-dimensional and stresses both preprocessing and fusion.

\paragraph{Steel Plates Faults.}
The Steel Plates Faults dataset contains $N=1941$ instances described by $d=27$ features and labels for
seven fault categories \cite{uci_steel}. The repository provides targets in a one-hot/multi-label form; we
convert them to a single class index via $\arg\max$ and retain all seven classes ($C=7$).

\paragraph{Cross-validation, monitor split, and leakage control.}
All reported results are obtained with stratified $K$-fold cross-validation.
Let $n_{c}$ denote the number of examples in class $c$; we cap $K$ by $\min\{5,\min_{c} n_{c}\}$ so that every fold
contains at least one example from each class. Within each outer training fold we optionally reserve a small
\emph{monitor} split (10\% of the fold training data by default) used exclusively for early stopping and model
selection; the monitor split is stratified and is never used to fit preprocessing operators or model parameters.

\subsection{Datasets and Experimental Setup}

We chose four benchmark datasets commonly used in hybrid learning evaluations: Wine, Breast Cancer, Forest CoverType, and a subset of Fashion-MNIST. Each dataset underwent standard preprocessing. For datasets with high input dimensionality (e.g., Fashion-MNIST), principal component analysis was applied to limit dimensionality to a manageable number of features.

We used stratified 5-fold cross-validation across all experiments to ensure fair and reproducible evaluation. To ensure determinism, random seed controls were applied across NumPy, PyTorch, scikit-learn, and other stochastic sources.

For each fold, models were trained using binary cross-entropy loss for binary classification tasks (e.g., Breast Cancer) and multiclass cross-entropy where applicable. The Adam optimizer was used with a OneCycle learning rate scheduler. All training and evaluation code is implemented in PyTorch and PennyLane.

\begin{figure}[H]
\centering
\begin{tikzpicture}[
    node distance=1.2cm,
    every node/.style={font=\footnotesize},
    enc/.style={
        rectangle, rounded corners, draw,
        minimum width=2.8cm,
        minimum height=1.0cm,
        align=center
    },
    small/.style={
        rectangle, rounded corners, draw,
        minimum width=2.3cm,
        minimum height=0.8cm,
        align=center
    },
    arrow/.style={->, thick}
]

\definecolor{classical}{RGB}{220,80,70}
\definecolor{quantum}{RGB}{240,180,60}
\definecolor{vqc}{RGB}{130,190,110}
\definecolor{measure}{RGB}{130,160,210}
\definecolor{fusion}{RGB}{160,120,200}
\definecolor{classifier}{RGB}{70,70,70}


\node[enc, draw=quantum] (qa_in) {Quantum Embedding\\ $|\psi_0(\mathbf{x})\rangle$};

\node[enc, draw=vqc, below=1.3cm of qa_in] (qa_v1)
{Variational Layer 1\\ $U_{\phi}^{(1)}$};

\node[enc, draw=vqc, below=1.0cm of qa_v1] (qa_v2)
{Variational Layer 2\\ $U_{\phi}^{(2)}$};

\node[enc, draw=measure, below=1.0cm of qa_v2] (qa_meas)
{Measurement\\ $\mathbf{z}_{\mathrm{qm}}$};

\node[enc, draw=classifier, below=1.0cm of qa_meas] (qa_clf)
{Classifier};

\draw[arrow] (qa_in) -- (qa_v1);
\draw[arrow] (qa_v1) -- (qa_v2);
\draw[arrow] (qa_v2) -- (qa_meas);
\draw[arrow] (qa_meas) -- (qa_clf);

\node[below=0.2cm of qa_clf] {(a) Pure QML};


\node[enc, draw=quantum, right=5cm of qa_in] (hb_in)
{Quantum Embedding\\ $|\psi_0(\mathbf{x})\rangle$};

\node[enc, draw=vqc, below=1.3cm of hb_in] (hb_v1)
{Variational Circuit\\ $U_{\phi}$};

\node[enc, draw=measure, below=1.0cm of hb_v1] (hb_meas)
{Measurement\\ $\mathbf{h}_{\mathrm{qm}}$};

\node[enc, draw=classical, below=1.0cm of hb_meas] (hb_clenc)
{Classical MLP\\ $g_{\theta}(\mathbf{h}_{\mathrm{qm}})$};

\draw[arrow] (hb_in) -- (hb_v1);
\draw[arrow] (hb_v1) -- (hb_meas);
\draw[arrow] (hb_meas) -- (hb_clenc);

\node[below=0.2cm of hb_clenc] {(b) Classical-on-Quantum (Original Hybrid)};


\node[enc, draw=classical, below=2.4cm of $(qa_clf)!0.2!(hb_clenc)$] (mf_clenc)
{Classical Encoder\\ $\mathbf{h}_{\mathrm{cl}} = f_{\theta}(\mathbf{x})$};

\node[enc, draw=quantum, right=4cm of mf_clenc] (mf_qemb)
{Quantum Embedding\\ $|\psi_0(\mathbf{x})\rangle$};

\node[enc, draw=vqc, below=1.2cm of mf_qemb] (mf_vqc)
{Variational Circuit\\ $U_{\phi}$};

\node[enc, draw=measure, below=1cm of mf_vqc] (mf_meas)
{Measurement\\ $\mathbf{h}_{\mathrm{qm}}$};

\draw[arrow] (mf_qemb) -- (mf_vqc);
\draw[arrow] (mf_vqc) -- (mf_meas);

\node[enc, draw=fusion, below=1.4cm of mf_clenc] (mf_att)
{Cross-Attention Fusion};

\draw[arrow] (mf_clenc) -- (mf_att);
\draw[arrow] (mf_meas.west) -- ++(-1.0cm,0) |- (mf_att.east);

\node[enc, draw=classifier, below=1.4cm of mf_att] (mf_clf)
{Classifier};

\draw[arrow] (mf_att) -- (mf_clf);

\node[below=0.2cm of mf_clf] {(c) Our Mid-Fusion HQC};

\end{tikzpicture}

\caption{
Model comparison: (a) pure quantum model using variational circuits only; (b) original hybrid QML approach where a classical MLP is trained on quantum measurement outputs; (c) our proposed multimodal hybrid QML model using cross-attention fusion between classical and quantum representations.}
\label{fig:model-comparison}
\end{figure}

\subsection{Problem Setting and Notation}
We consider supervised classification on a tabular dataset
$\mathcal{D}=\{(\bm{x}_i, y_i)\}_{i=1}^{N}$, where $\bm{x}_i\in\mathbb{R}^{d}$ is a real-valued feature vector,
$N$ is the number of instances, $d$ is the raw feature dimension, and
$y_i\in\{0,\dots,C-1\}$ is a discrete class label (with $C=2$ for binary tasks, i.e., $y_i\in\{0,1\}$).
Bold symbols denote vectors (e.g., $\bm{x}$), and non-bold symbols denote scalars (e.g., $a$, $\alpha$).
The learning objective is to construct a predictor
$f(\bm{x})$ that outputs class probabilities or logits (pre-sigmoid / pre-softmax scores) used for decision-making.

The benchmark evaluates a family of \emph{classical}, \emph{quantum}, and \emph{hybrid fusion} models.
All methods share a common preprocessing pipeline and differ only in how classical representations and
quantum circuit readouts are formed and combined.
For clarity, we write
\begin{equation}
\bm{x}^{(c)} = \Phi_c(\bm{x}), \qquad \bm{x}^{(q)} = \Phi_q(\bm{x}),
\end{equation}
for the (possibly different) inputs to the classical and quantum branches, respectively, where $\Phi_c$ and
$\Phi_q$ denote the branch-specific transforms (standardization and optional PCA, described below).
In the final benchmark regime the same feature vector is provided to both branches, i.e.,
$\bm{x}^{(c)}=\bm{x}^{(q)}=\bm{x}$ up to the branch-specific transforms described below.
This design avoids confounding fusion effects with an arbitrary partition of features.

\section{Preprocessing and Dimensionality Control}
\subsection{Standardization}
Each fold is standardized using the mean and standard deviation estimated on the corresponding training split,
\begin{equation}
\tilde{\bm{x}} = \frac{\bm{x} - \bm{\mu}}{\bm{\sigma}},
\end{equation}
where $\bm{\mu}\in\mathbb{R}^{d}$ and $\bm{\sigma}\in\mathbb{R}^{d}$ are the per-feature training mean and
standard deviation (division is element-wise), and the same $(\bm{\mu},\bm{\sigma})$ are applied to the fold test split.
(Here $\bm{\sigma}$ denotes standard deviation; later $\sigma(\cdot)$ denotes the logistic sigmoid function.)
This ensures numerical comparability across features and stabilizes both
classical optimization and angle encoding in the quantum branch.

\subsubsection{Principal Component Projections}
Tabular datasets often have feature dimension $d$ larger than the number of available qubits.
To keep the quantum circuit width fixed while limiting information loss, we apply PCA~\cite{jolliffe2002pca}
to the standardized features \emph{within each fold} (fit on training data only).
\begin{itemize}
  \item \textbf{Quantum projection:} a PCA map $\bm{x}^{(q)} = \bm{P}_q^\top \tilde{\bm{x}}$.
  We retain $Q=\min(Q_{\max}, d, n_{\mathrm{train}})$ components, where $n_{\mathrm{train}}$ is the number of training
  samples in the current fold and $Q_{\max}=9$ in the final setting.
  This caps the quantum input dimension (and hence the number of qubits) to the available qubit budget.
  \item \textbf{Classical projection (optional):} a PCA map
  $\bm{x}^{(c)} = \bm{P}_c^\top \tilde{\bm{x}}$ retaining $95\%$ of the training variance by default.
  Several baselines intentionally disable this step and use the full standardized feature vector;
  this is reported per model definition when relevant.
\end{itemize}
The use of fold-specific transforms prevents test leakage and keeps comparisons consistent across models.

\subsection{Quantum Branch: Variational Circuit and Readout}
\subsubsection{Angle Encoding with Trainable Input Scaling}
Let $\bm{x}^{(q)}\in\mathbb{R}^{Q}$ denote the quantum-branch features after \emph{standardization and PCA}.
The circuit uses angle encoding via single-qubit $Y$ rotations.
To reduce aliasing from the $2\pi$ periodicity of rotations while retaining differentiability,
the angles are bounded as
\begin{equation}
\bm{\theta} = \pi\, \tanh(\bm{s}\odot \bm{x}^{(q)}),
\label{eq:angle_scaling}
\end{equation}
where $\bm{s}\in\mathbb{R}^{Q}$ is a trainable per-wire scale vector, $\tanh(\cdot)$ is applied element-wise, and $\odot$ is the Hadamard product.
Trainable scaling is a lightweight mechanism to adjust the effective frequency of the embedding and can
improve optimization in practice, especially when input ranges vary across datasets~\cite{mitarai2018qcl, cerezo2021vqa}.

\subsubsection{Variational Ansatz}
Starting from $\lvert 0\rangle^{\otimes Q}$, the embedding applies
$R_y(\theta_j)$ on wire $j$, followed by $L$ layers of a hardware-efficient entangling template
implemented as \emph{Strongly Entangling Layers}~\cite{bergholm2018pennylane,PQCs2019_Benedetti}.
The variational parameters can be collected into a tensor
$\bm{W}\in\mathbb{R}^{L\times Q\times 3}$, corresponding to three Euler angles per qubit per layer.
Denoting the overall unitary by $U(\bm{\theta},\bm{W})$, the resulting quantum state is
\begin{equation}
\lvert \psi(\bm{x}^{(q)};\bm{W})\rangle = U(\bm{\theta},\bm{W})\lvert 0\rangle^{\otimes Q}.
\end{equation}
In the final benchmark, the circuit width and depth are fixed to $Q=9$ qubits and $L=3$ variational layers.
Weights are initialized in a small neighborhood of zero to reduce the risk of flat loss landscapes
in shallow variational models~\cite{McClean2018Barren}.

\subsubsection{Observable Readout}
The quantum circuit is converted into a real-valued feature vector via expectation values.
Let $\lvert\psi\rangle \equiv \lvert \psi(\bm{x}^{(q)};\bm{W})\rangle$ for brevity.
The default readout used throughout the final experiments combines local Pauli-$Z$ statistics with
nearest-neighbor $ZZ$ correlations on a ring:
\begin{align}
z_j(\bm{x}^{(q)}) &= \langle \psi \rvert Z_j \lvert \psi \rangle, \qquad j=0,\dots,Q-1,\\
z_{Q+j}(\bm{x}^{(q)}) &= \langle \psi \rvert Z_j Z_{j+1 \bmod Q} \lvert \psi \rangle,
\qquad j=0,\dots,Q-1,
\label{eq:readout}
\end{align}
where $(j+1 \bmod Q)$ indicates periodic (ring) indexing.
This yields a $2Q$-dimensional quantum feature vector $\bm{z}\in\mathbb{R}^{2Q}$ that captures both
marginal and pairwise statistics in the learned quantum state.
Alternative readouts (e.g., $\{Z,X,Y\}$ per qubit or only ring correlations) can be defined similarly,
but the reported benchmark uses the combined readout for a consistent comparison.

Gradients of circuit parameters are obtained through differentiable simulation of expectation values,
using analytic differentiation rules supported by the simulator backend~\cite{bergholm2018pennylane}.
This enables end-to-end training of hybrid models in the same optimization loop as classical layers.

\subsection{Classical Branch: Neural Predictors on Tabular Features}
Classical components are implemented as multi-layer perceptrons (MLPs) with
GELU (Gaussian Error Linear Unit) nonlinearity~\cite{hendrycks2016gelu} and dropout regularization.
We denote a generic classical mapping by
\begin{equation}
\bm{h}_c = g_c(\bm{x}^{(c)};\bm{\phi}), \qquad \bm{h}_c\in\mathbb{R}^{D},
\end{equation}
where $D$ is a shared latent dimension (set to $D=64$ in the benchmark) and $\bm{\phi}$ denotes trainable classical parameters.
Unless stated otherwise, MLP blocks use dropout with probability $p=0.10$.
For attention-based fusion, the Transformer block uses $H=4$ attention heads and a feed-forward expansion factor of $4$.
Depending on the model, $g_c$ ranges from shallow two-block MLPs to deeper stacks with additional normalization (e.g., LayerNorm).
The strongest classical baseline uses a deeper MLP with LayerNorm and dropout, intended to approximate a
well-tuned ``best classical'' setting under the same training protocol.

\subsection{Model Families}
All models output logits $\bm{\ell}\in\mathbb{R}^{C}$ for multi-class tasks or $\ell\in\mathbb{R}$ for binary tasks.
For binary classification, probabilities are $p(y{=}1\mid \bm{x})=\sigma(\ell)$, while for multi-class tasks
$p(y{=}c\mid \bm{x})=\mathrm{softmax}(\bm{\ell})_c$, where $\sigma(t)=1/(1+e^{-t})$ is the logistic sigmoid.

\subsubsection{Classical-Only Baselines}
The classical-only predictor is a standard MLP,
\begin{equation}
\bm{\ell} = W_o\, g_c(\bm{x}^{(c)};\bm{\phi}) + \bm{b}_o.
\end{equation}
A deeper variant increases depth to test whether gains from hybrid models persist under stronger classical capacity.

\subsubsection{Quantum-Only Baselines}
Quantum-only models act on quantum features $\bm{x}^{(q)}$ through the variational circuit and a classical head:
\begin{equation}
\bm{z} = q(\bm{x}^{(q)};\bm{W}), \qquad \bm{\ell} = W_o\, \bm{z} + \bm{b}_o,
\end{equation}
where $q(\cdot)$ denotes the mapping induced by~\eqref{eq:angle_scaling}--\eqref{eq:readout}.
A stronger quantum baseline additionally replaces the linear head by a deeper MLP operating on $\bm{z}$,
mirroring the strongest classical baseline in spirit.

\subsubsection{Early Fusion}
Early fusion concatenates the classical features with the quantum readout features before classification:
\begin{equation}
\bm{u} = \big[\bm{x}^{(c)}; \bm{z}\big], \qquad \bm{\ell} = g_e(\bm{u};\bm{\eta}),
\end{equation}
where $[\cdot;\cdot]$ denotes concatenation and $g_e$ is an MLP with parameters $\bm{\eta}$.
This design exposes the classifier to both representations directly.
In this benchmark, early fusion uses the standardized raw classical features (no classical PCA) so that
fusion is not limited by an information bottleneck imposed by dimensionality reduction.

\subsubsection{Late Fusion (Logit Mixing)}
Late fusion combines separate classical and quantum predictors at the logit level:
\begin{equation}
\bm{\ell}_c = f_c(\bm{x}^{(c)}), \qquad \bm{\ell}_q = f_q(\bm{x}^{(q)}),
\end{equation}
\begin{equation}
\bm{\ell} = \alpha\, \bm{\ell}_c + (1-\alpha)\, \bm{\ell}_q,\qquad \alpha=\sigma(a),
\label{eq:latefusion}
\end{equation}
where $a\in\mathbb{R}$ is a learned scalar and $\sigma(\cdot)$ is the logistic sigmoid.
A deep late-fusion variant replaces the classical head $f_c$ by a deeper MLP while keeping the same mixing rule.
Because the mixture is applied to logits, the combination is task-aligned and does not require the two branches
to share a latent space.

\subsubsection{Mid-Level Fusion by Latent Mixing}
Mid-level fusion first maps each branch into a shared latent space $\mathbb{R}^{D}$ and then mixes features:
\begin{equation}
\bm{h}_c = W_c\, \bm{x}^{(c)} + \bm{b}_c,\qquad
\bm{h}_q = W_q\, \bm{z} + \bm{b}_q,
\end{equation}
\begin{equation}
\bm{h} = \alpha\, \bm{h}_c + (1-\alpha)\, \bm{h}_q,\qquad \alpha=\sigma(a),
\end{equation}
\begin{equation}
\bm{\ell} = W_o\, \bm{h} + \bm{b}_o.
\label{eq:midfusion}
\end{equation}
Compared with~\eqref{eq:latefusion}, the mixing occurs before the final classifier, allowing the model to form
a shared representation that can be linearly separable even when each branch alone is insufficient.

\subsubsection{Token-Based Mid Fusion with Self-Attention}
To allow richer interactions between individual quantum measurements and classical features, we define a
sequence model where the classical representation forms a ``CLS'' (classification) token and each quantum readout component is
treated as a token.
Let $\bm{z}\in\mathbb{R}^{M}$ with $M=2Q$.
Each scalar $z_m$ is embedded as $\bm{t}_m = W_t z_m + \bm{b}_t$ (for $m=1,\dots,M$) and augmented with a learned identity embedding
$\bm{e}_m$ to disambiguate measurement indices.
The classical token is $\bm{t}_0 = W_c \bm{x}^{(c)} + \bm{b}_c$.
We obtain the sequence
\begin{equation}
\bm{T} = [\bm{t}_0;\, \bm{t}_1+\bm{e}_1;\, \dots;\, \bm{t}_M+\bm{e}_M] \in \mathbb{R}^{(M+1)\times D}.
\end{equation}
A Transformer-style self-attention block~\cite{vaswani2017attention} produces contextualized tokens
$\bm{T}'=\mathrm{Attn}(\bm{T})$, after which the classifier reads out the first (CLS) token:
\begin{equation}
\bm{\ell} = W_o\, \bm{T}'_{0} + \bm{b}_o.
\end{equation}
A deep classical fusion variant replaces the initial classical projection by a deeper MLP, retaining the same
tokenization and attention mechanism. This isolates the effect of increasing classical expressivity from
the effect of using attention for cross-branch interaction.

\subsubsection{Deep Fusion and Very Deep Fusion}
For deep fusion variants, each branch first applies a deeper feature extractor before gated latent mixing.
Fusion then follows~\eqref{eq:midfusion}.
Two settings are studied: $k=3$ (deep fusion) and $k=4$ (very deep fusion), where $k$ denotes the depth of the classical MLP trunk used before fusion.

\subsection{Optimization and Evaluation Protocol}
\subsubsection{Cross-Validation and Early Stopping}
Performance is estimated using stratified $K$-fold cross-validation ($K=5$), ensuring class proportions are
preserved across folds.
Within each training fold, an inner stratified monitor split (10\% of the fold training set, when feasible) is
used for early stopping.
Training stops when the monitored quantity fails to improve by at least $10^{-4}$ for a patience of 7 epochs.
The default monitor target is macro-F1; an alternative condition monitors validation loss to test sensitivity
to the stopping criterion.

\subsubsection{Loss Functions}
For binary classification, logits $\ell$ are trained with the sigmoid cross-entropy loss:
\begin{equation}
\mathcal{L}_{\text{bin}} = -\frac{1}{N}\sum_{i=1}^{N}\Big[y_i\log\sigma(\ell_i) + (1-y_i)\log(1-\sigma(\ell_i))\Big].
\end{equation}
For multi-class classification, the loss is the cross-entropy with label smoothing~\cite{szegedy2016labelsmoothing}
(smoothing parameter $\varepsilon=0.05$ in the reported experiments).
Let $\bm{p}_i=\mathrm{softmax}(\bm{\ell}_i)$ and $\bm{y}_i$ be the one-hot label vector.
With smoothing parameter $\varepsilon$, the smoothed target distribution can be written as
$\bm{y}_i^{(\varepsilon)}=(1-\varepsilon)\bm{y}_i + (\varepsilon/C)\mathbf{1}$, equivalently (component-wise)
$y_{ic}^{(\varepsilon)}=(1-\varepsilon)y_{ic}+\varepsilon/C$, and
\begin{equation}
\mathcal{L}_{\text{mc}} = -\frac{1}{N}\sum_{i=1}^{N}\sum_{c=1}^{C} y_{ic}^{(\varepsilon)} \log p_{ic}.
\end{equation}

\subsubsection{Optimization Details}
All models are trained with AdamW~\cite{loshchilov2019adamw} (Adam with decoupled weight decay) for up to 30 epochs using mini-batches of size 64.
The base learning rate is $10^{-3}$ with weight decay $10^{-3}$.
Gradients are clipped to unit norm (max-norm 1.0) to avoid unstable updates.
A warmup-and-decay learning-rate schedule is applied, consisting of a linear warmup over the first 10\% of
optimizer steps followed by cosine decay to a minimum learning rate equal to 10\% of the base rate.
These choices are standard in both deep learning and variational quantum model training, and they help
equalize optimization conditions across model families.

\subsubsection{Metrics}
On each test fold, we report accuracy and macro-averaged precision, recall, and F1.
Macro averaging assigns equal weight to each class and is therefore informative under class imbalance.
ROC-AUC (area under the receiver operating characteristic curve) is computed as standard AUC for binary tasks and as an average one-vs-rest (OvR) AUC for multi-class tasks when
the metric is well-defined; folds lacking the required class support are treated as undefined and excluded
from the corresponding mean via NaN-safe aggregation.

\subsection{Computational Considerations}
Quantum simulation cost scales exponentially in the number of qubits, $\mathcal{O}(2^Q)$ in a statevector model.
Fixing $Q=9$ keeps simulation tractable while still enabling non-trivial entanglement and correlation readouts.
The combined readout in~\eqref{eq:readout} returns $2Q$ expectation values per circuit evaluation, which are then
treated as learnable features inside the hybrid networks. This balance between circuit expressivity and tractability
is aligned with common practice in benchmarking variational classifiers on classical data~\cite{cerezo2021vqa,benedetti2019qnn}.

\section{Results}
\label{sec:results}

\begin{figure}[!t]
  \centering
  \includegraphics[width=\textwidth]{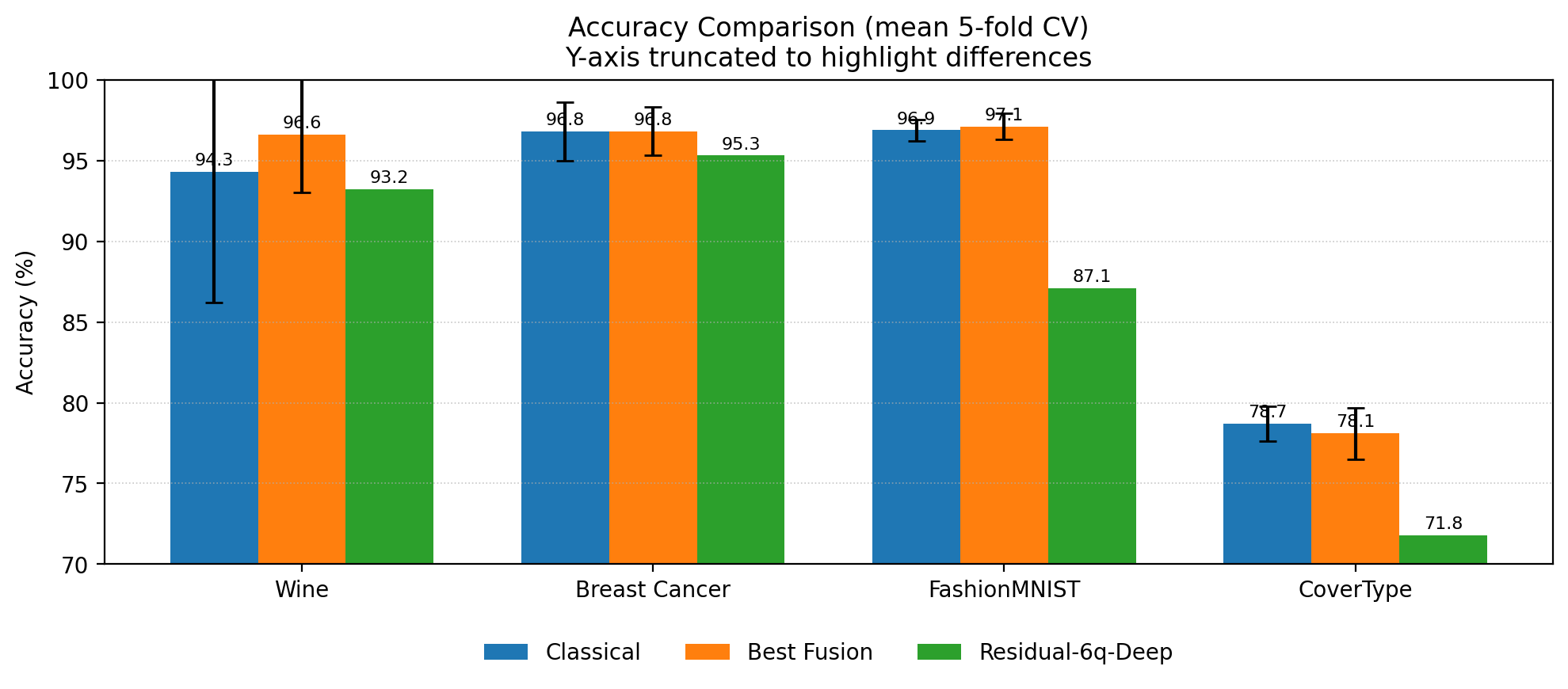}
  \caption{Accuracy comparison across datasets (mean over 5-fold cross-validation). We report Classical, our best-performing Fusion variant per dataset (“Best Fusion”), and the Residual-6q-Deep baseline. Error bars denote the standard deviation across folds where available (Classical and Best Fusion); Residual-6q-Deep reports mean accuracy only. The y-axis is truncated to the 70--100\% range to make high-accuracy differences visible; the Quantum-only baseline is omitted from this plot to avoid compressing the scale.}
  \label{fig:acc_bar_zoom}
\end{figure}

We summarize five-fold cross-validation test performance for each dataset. Reported values are mean $\pm$ standard deviation (std) over folds for Accuracy, Precision, Recall, F1, and ROC-AUC. The primary selection criterion is the F1 score; the highest F1 in each table is highlighted in bold.

As shown in Fig.~\ref{fig:acc_bar_zoom}, our best fusion configuration matches or exceeds the classical baseline on Wine, Breast Cancer, and FashionMNIST, while remaining competitive on CoverType. To emphasize practically relevant differences in the high-accuracy regime, the plot uses a truncated y-axis and omits the quantum-only baseline, which otherwise compresses the scale.

\begin{table}[!t]
\centering
\caption{Accuracy (\%) comparison across datasets. Bold indicates the highest accuracy per dataset (row-wise). In case of a tie, we bold \emph{Best Fusion} to keep a single highlighted value per row. Residual-6q-Deep does not report Steel and is shown as --.}
\label{tab:acc_summary}
\setlength{\tabcolsep}{10pt}
\renewcommand{\arraystretch}{1.15}
\begin{tabular}{lcccc}
\toprule
Dataset & Classical & Quantum & Best Fusion & Residual-6q-Deep \\
\midrule
Wine          & 94.3 & 30.8 & \textbf{96.6} & 93.2 \\
Breast Cancer & 96.8 & 49.0 & \textbf{96.8} & 95.3 \\
FashionMNIST  & 96.9 & 49.4 & \textbf{97.1} & 87.1 \\
CoverType     & \textbf{78.7} & 49.0 & 78.1 & 71.8 \\
Steel         & \textbf{75.1} & 45.8 & 74.9 & \multicolumn{1}{c}{--} \\
\bottomrule
\end{tabular}
\end{table}

Table.~\ref{tab:acc_summary} summarizes the accuracy results, including Steel. We highlight the best reported value in each method column to provide an at-a-glance view of each method’s peak performance across the benchmark suite.

\paragraph{Wine.} Table~\ref{tab:wine_results} shows that \texttt{midfusion\_attn(ours)} achieves the best F1 on Wine (\num{0.9671 +- 0.0359}) with near-ceiling ROC-AUC (\num{0.9986 +- 0.0021}). Relative to the classical baseline (\texttt{best\_classical}, F1=\num{0.9434 +- 0.0792}), the strongest fusion variant yields a substantial improvement, while the quantum-only model underperforms (F1=\num{0.2052 +- 0.0168}).

\paragraph{BreastCancer.} On BreastCancer (Table~\ref{tab:breastcancer_results}), the classical baseline \texttt{best\_classical} attains the best F1 (\num{0.9665 +- 0.0189}), with \texttt{midfusion\_attn(ours)} closely matching it. Quantum-only performance is markedly lower (F1=\num{0.4052 +- 0.1015}).

\paragraph{CoverType.} For CoverType (Table~\ref{tab:covertype_results}), \texttt{best\_classical} provides the highest F1 (\num{0.8106 +- 0.0093}), and the best fusion configuration (\texttt{early\_fusion(ours)}) is competitive (F1=\num{0.8063 +- 0.0181}). The quantum-only model again trails substantially (F1=\num{0.3351 +- 0.0363}).

\paragraph{FashionMNIST.} On FashionMNIST (Table~\ref{tab:fashionmnist_results}), \texttt{early\_fusion(ours)} achieves the best F1 (\num{0.9710 +- 0.0079}), slightly outperforming the classical baseline (F1=\num{0.9687 +- 0.0068}).

\paragraph{SteelPlatesFaults} On SteelPlatesFaults (Table~\ref{tab:steelplatesfaults_results}), \texttt{early\_fusion(ours)} achieves the best F1 (\num{0.9710 +- 0.0079}), slightly outperforming the classical baseline (F1=\num{0.9687 +- 0.0068}).

\begin{table}[H]
\centering
\small
\caption{Five-fold test performance on \textbf{Wine} (mean $\pm$ std over folds). Best result is shown in bold.}
\label{tab:wine_results}
\sisetup{table-number-alignment=center,table-text-alignment=center}
\begin{tabular}{l S[table-format=1.4, separate-uncertainty=true] S[table-format=1.4, separate-uncertainty=true] S[table-format=1.4, separate-uncertainty=true] S[table-format=1.4, separate-uncertainty=true] S[table-format=1.4, separate-uncertainty=true]}
\toprule
Model & {Accuracy} & {Precision} & {Recall} & {F1} & {ROC-AUC} \\
\midrule
\texttt{best\_classical} & \num{0.943 +- 0.081} & \num{0.945 +- 0.075} & \num{0.950 +- 0.068} & \num{0.943 +- 0.079} & \num{0.982 +- 0.035} \\
\texttt{quantum\_only} & \num{0.308 +- 0.032} & \num{0.246 +- 0.119} & \num{0.318 +- 0.033} & \num{0.205 +- 0.017} & \num{0.452 +- 0.069} \\
\texttt{midfusion\_linear(ours)} & \num{0.842 +- 0.048} & \num{0.852 +- 0.042} & \num{0.857 +- 0.043} & \num{0.844 +- 0.047} & \num{0.956 +- 0.018} \\
\texttt{midfusion\_attn(ours)} & {\bfseries \num{0.966 +- 0.036}} & {\bfseries \num{0.968 +- 0.035}} & {\bfseries \num{0.970 +- 0.033}} & {\bfseries \num{0.967 +- 0.036}} & {\bfseries \num{0.999 +- 0.002}} \\
\texttt{early\_fusion(ours)} & \num{0.927 +- 0.032} & \num{0.930 +- 0.029} & \num{0.938 +- 0.028} & \num{0.928 +- 0.033} & \num{0.990 +- 0.007} \\
\texttt{late\_fusion(ours)} & \num{0.860 +- 0.068} & \num{0.882 +- 0.059} & \num{0.855 +- 0.081} & \num{0.852 +- 0.077} & \num{0.964 +- 0.026} \\

\bottomrule
\end{tabular}
\end{table}

\begin{table}[H]
\centering
\small
\caption{Five-fold test performance on \textbf{BreastCancer} (mean $\pm$ std over folds). Best result is shown in bold.}
\label{tab:breastcancer_results}
\sisetup{table-number-alignment=center,table-text-alignment=center}
\begin{tabular}{l S[table-format=1.4, separate-uncertainty=true] S[table-format=1.4, separate-uncertainty=true] S[table-format=1.4, separate-uncertainty=true] S[table-format=1.4, separate-uncertainty=true] S[table-format=1.4, separate-uncertainty=true]}
\toprule
Model & {Accuracy} & {Precision} & {Recall} & {F1} & {ROC-AUC} \\
\midrule
\texttt{best\_classical} & {\bfseries \num{0.968 +- 0.018}} & {\bfseries \num{0.967 +- 0.024}} & {\bfseries \num{0.968 +- 0.013}} & {\bfseries \num{0.966 +- 0.019}} &  \num{0.992 +- 0.009} \\
\texttt{quantum\_only} & \num{0.490 +- 0.119} & \num{0.457 +- 0.165} & \num{0.503 +- 0.034} & \num{0.405 +- 0.101} & \num{0.502 +- 0.046} \\
\texttt{midfusion\_linear(ours)} & \num{0.954 +- 0.024} & \num{0.960 +- 0.023} & \num{0.944 +- 0.030} & \num{0.950 +- 0.027} & \num{0.990 +- 0.008} \\
\texttt{midfusion\_attn(ours)} & {\bfseries \num{ 0.968 +- 0.015}} & {\bfseries \num{0.967 +- 0.018}} &  \num{0.967 +- 0.016} & {\bfseries \num{0.966 +- 0.016}} & {\bfseries \num{0.995 +- 0.005}} \\
\texttt{early\_fusion(ours)} & \num{0.963 +- 0.023} & \num{0.969 +- 0.026} & \num{0.958 +- 0.023} & \num{0.960 +- 0.024} & \num{0.994 +- 0.004} \\
\texttt{late\_fusion(ours)} & \num{0.958 +- 0.024} & \num{0.9614 +- 0.025} & \num{0.949 +- 0.028} & \num{0.9542 +- 0.0265} & \num{0.9897 +- 0.009} \\
\bottomrule
\end{tabular}
\end{table}

\begin{table}[H]
\centering
\small
\caption{Five-fold test performance on \textbf{CoverType} (mean $\pm$ std over folds). Best result is shown in bold.}
\label{tab:covertype_results}
\sisetup{table-number-alignment=center,table-text-alignment=center}
\begin{tabular}{l S[table-format=1.4, separate-uncertainty=true] S[table-format=1.4, separate-uncertainty=true] S[table-format=1.4, separate-uncertainty=true] S[table-format=1.4, separate-uncertainty=true] S[table-format=1.4, separate-uncertainty=true]}
\toprule
Model & {Accuracy} & {Precision} & {Recall} & {F1} & {ROC-AUC} \\
\midrule
\texttt{best\_classical} & {\bfseries\num{0.787 +- 0.011}} & {\bfseries \num{0.805 +- 0.010}} & {\bfseries \num{0.818 +- 0.014}} & {\bfseries \num{0.811 +- 0.009}} & \num{0.914 +- 0.007} \\
\texttt{quantum\_only} & \num{0.490 +- 0.018} & \num{0.455 +- 0.141} & \num{0.357 +- 0.019} & \num{0.335 +- 0.036} & \num{0.539 +- 0.024} \\
\texttt{midfusion\_linear(ours)} & \num{0.742 +- 0.013} & \num{0.761 +- 0.018} & \num{0.785 +- 0.011} & \num{0.771 +- 0.013} & \num{0.887 +- 0.009} \\
\texttt{midfusion\_attn(ours)} & \num{0.762 +- 0.027} & \num{0.778 +- 0.036} & \num{0.797 +- 0.024} & \num{0.780 +- 0.028} & \num{0.899 +- 0.013} \\
\texttt{early\_fusion(ours)} & \num{0.781 +- 0.016} & \num{0.802 +- 0.019} & \num{0.812 +- 0.018} & \num{0.806 +- 0.018} & {\bfseries\num{0.914 +- 0.008}} \\
\texttt{late\_fusion(ours)} & \num{0.769 +- 0.019} & \num{0.786 +- 0.026} & \num{0.803 +- 0.013} & \num{0.794 +- 0.020} & \num{0.900 +- 0.010} \\
\bottomrule
\end{tabular}
\end{table}
.
\begin{table}[t]
\centering
\small
\caption{Five-fold test performance on \textbf{FashionMNIST} (mean $\pm$ std over folds). Best result is shown in bold.}
\label{tab:fashionmnist_results}
\sisetup{table-number-alignment=center,table-text-alignment=center}
\begin{tabular}{l S[table-format=1.4, separate-uncertainty=true] S[table-format=1.4, separate-uncertainty=true] S[table-format=1.4, separate-uncertainty=true] S[table-format=1.4, separate-uncertainty=true] S[table-format=1.4, separate-uncertainty=true]}
\toprule
Model & {Accuracy} & {Precision} & {Recall} & {F1} & {ROC-AUC} \\
\midrule
\texttt{best\_classical} & \num{0.969 +- 0.007} & \num{0.969 +- 0.007} & \num{0.969 +- 0.007} & \num{0.969 +- 0.007} & \num{0.994 +- 0.003} \\
\texttt{quantum\_only} & \num{0.494 +- 0.032} & \num{0.49 +- 0.03} & \num{0.49 +- 0.03} & \num{0.491 +- 0.032} & \num{0.686 +- 0.030} \\
\texttt{midfusion\_linear(ours)} & \num{0.963 +- 0.011} & \num{0.964 +- 0.012} & \num{0.963 +- 0.011} & \num{0.963 +- 0.012} & \num{0.992 +- 0.002} \\
\texttt{midfusion\_attn(ours)} & \num{0.963 +- 0.012} & \num{0.963 +- 0.012} & \num{0.963 +- 0.012} & \num{0.963 +- 0.012} & \num{0.995 +- 0.002} \\
\texttt{early\_fusion(ours)} &  {\bfseries \num{ 0.971 +- 0.008}} &  {\bfseries \num{ 0.971 +- 0.008}} &  {\bfseries \num{ 0.971 +- 0.008}} & {\bfseries \num{0.971 +- 0.008}} &  {\bfseries \num{ 0.994 +- 0.003}} \\
\texttt{late\_fusion(ours)} & \num{0.964 +- 0.007} & \num{0.964 +- 0.006} & \num{0.964 +- 0.007} & \num{0.964 +- 0.007} & \num{0.994 +- 0.004} \\
\bottomrule
\end{tabular}
\end{table}

\begin{table}[H]
\centering
\small
\caption{Five-fold test performance on \textbf{SteelPlatesFaults} (mean $\pm$ std over folds). Best mean F1 is shown in bold.}
\label{tab:steelplatesfaults_results}
\sisetup{table-number-alignment=center,table-text-alignment=center}
\begin{tabular}{l S[table-format=1.4, separate-uncertainty=true] S[table-format=1.4, separate-uncertainty=true] S[table-format=1.4, separate-uncertainty=true] S[table-format=1.4, separate-uncertainty=true] S[table-format=1.4, separate-uncertainty=true]}
\toprule
Model & {Accuracy} & {Precision} & {Recall} & {F1} & {ROC-AUC} \\
\midrule
\texttt{best\_classical} & {\bfseries \num{ 0.751 +- 0.018}} & \num{0.768 +- 0.005} & {\bfseries \num{ 0.774 +- 0.033}} &  \num{ 0.768 +- 0.015} & {\bfseries \num{0.948 +- 0.005}} \\
\texttt{quantum\_only} & \num{0.458 +- 0.020} & \num{0.403 +- 0.047} & \num{0.375 +- 0.032} & \num{0.381 +- 0.038} & \num{0.740 +- 0.029} \\
\texttt{midfusion\_linear(ours)} & \num{0.694 +- 0.025} & \num{0.711 +- 0.051} & \num{0.624 +- 0.027} & \num{0.640 +- 0.034} & \num{0.917 +- 0.007} \\
\texttt{midfusion\_attn(ours)} & \num{0.749 +- 0.023} & {\bfseries \num{ 0.778 +- 0.012}} & \num{0.768 +- 0.028} & {\bfseries \num{0.771 +- 0.012}} & \num{0.942 +- 0.011} \\
\texttt{early\_fusion(ours)} & \num{0.729 +- 0.022} & \num{0.759 +- 0.019} & \num{0.726 +- 0.035} & \num{0.738 +- 0.025} & \num{0.939 +- 0.007} \\
\texttt{late\_fusion(ours)} & \num{0.730 +- 0.021} & \num{0.749 +- 0.053} & \num{0.705 +- 0.046} & \num{0.717 +- 0.047} & \num{0.939 +- 0.012} \\
\bottomrule
\end{tabular}
\end{table}

\section{Conclusion}
\label{sec:conclusion}

This work explored hybrid quantum--classical learning through a multimodal lens, treating quantum and classical representations not as competing alternatives but as complementary sources of information. By doing so, we were able to rethink how hybrid models should be constructed, moving beyond simple concatenation of quantum outputs toward fusion strategies that allow the two pathways to interact more deliberately.

Across four benchmark datasets, our experiments show that quantum circuits in isolation, and even in their conventional hybrid form, struggle to match classical baselines. As such, the results reinforce what has been increasingly recognised in the literature: quantum-enhanced learning is unlikely to outperform classical approaches unless the interaction between modalities is modelled carefully. Pure quantum models suffer most severely from the measurement bottleneck, and although classical post-processing partially alleviates this issue, it is not sufficient to close the gap with classical MLPs.

Our proposed cross-attention mid-fusion approach offers a more effective alternative. By letting the classical pathway query quantum-derived features selectively, MidFusion consistently performs best on complex datasets, improving upon both classical and hybrid baselines. These gains suggest that quantum encoders can provide useful global structure, but only when integrated through a mechanism that can adaptively determine when and how such information should be incorporated. In simpler datasets, where the classical decision boundary is already well supported by the data, MidFusion matches the performance of MLPs and other hybrids, indicating that it introduces no unnecessary overhead or instability when quantum features are redundant.

Thus, this study presents a more grounded view of quantum-assisted learning: the value of quantum circuits lies not in replacing classical components but in enriching them, particularly through structured fusion that respects the multimodal nature of the problem. By framing hybrid QML models in this way, we open a path toward architectures that leverage quantum structure without overclaiming advantage.

Future work may examine the role of deeper circuit families, error-mitigation techniques, or problem-aligned encodings in further enhancing the quantum branch of the model. In addition, multimodal strategies beyond attention, such as co-training, contrastive alignment, or uncertainty-aware fusion, may reveal further opportunities for meaningful quantum–classical interaction. As quantum hardware continues to mature, such hybrid and multimodal perspectives will likely become essential for designing models that make the most of both worlds: the classical richness of data-driven pattern recognition and the unique representational structure offered by quantum computation.

\paragraph{Declaration of generative AI and AI-assisted technologies in the manuscript preparation process:}

During the preparation of this work the authors used Open AI in order to get assistance with editing, and structuring the manuscript. After using this tool/service, the authors reviewed and edited the content as needed and takes full responsibility for the content of the published article.

\bibliographystyle{IEEEtran}
\bibliography{refs}

\end{document}